\documentclass[letterpaper, 10 pt, conference]{ieeeconf}  

\IEEEoverridecommandlockouts                              

\usepackage{graphics} 
\usepackage{graphicx}
\usepackage{subcaption}

\title{\LARGE \bf
Free-View, 3D Gaze-Guided, Assistive Robotic System \\for Activities of Daily Living
}

\author{Ming-Yao Wang$^{*}$, Alexandros A. Kogkas$^{*}$, Ara Darzi, and George P. Mylonas, \IEEEmembership{Member, IEEE}%
\thanks{*Ming-Yao Wang and Alexandros A. Kogkas are joint first authors.}
\thanks{All authors are with the HARMS Lab, Department of Surgery and Cancer, Imperial College London, W21PF London, UK {\tt\small (e-mail: a.kogkas15@imperial.ac.uk)}}%
}

\begin{document}

\maketitle
\thispagestyle{empty}
\pagestyle{empty}

\begin{abstract}

Patients suffering from quadriplegia have limited body motion which prevents them from performing daily activities. We have developed an assistive robotic system with an intuitive free-view gaze interface. The user's point of regard is estimated in 3D space while allowing free head movement and is combined with object recognition and trajectory planning. This framework allows the user to interact with objects using fixations. Two operational modes have been implemented to cater for different eventualities. The automatic mode performs a pre-defined task associated with a gaze-selected object, while the manual mode allows gaze control of the robot's end-effector position on the user's frame of reference. User studies reported effortless operation in automatic mode. A manual pick and place task achieved a success rate of 100\% on the users' first attempt.

\end{abstract}

\section{INTRODUCTION}

Quadriplegia is the partial or total paralysis of all four limbs. Various illness or injury can result in this condition such as cerebral palsy, amyotrophic lateral sclerosis (ALS), muscular dystrophy, traumatic brain or spinal injury and stroke. Being unable to move around or handle objects present difficult challenges to one's daily life. For many patients, the desire to regain mobility or at least dexterity so they do not feel completely helpless, is a longing wish.

\textit{"It would almost be easier if the arms came back. You could sit in a wheelchair, at least you could do something. When the leg comes back the only thing you learn to do is walk. But the number of things you can do with an arm..."} \cite{barker2005upper}.

Nowadays, there are wheelchair-mounted robotic manipulators (WMRM) available such as the JACO\footnotemark[1] or iARM\footnotemark[2] to allow these patients to gain dexterity. The arm can be manually controlled using a joystick and pushbuttons. However, this may not be possible for patients who suffer from severe motion disabilities.

Electroencephalography (EEG) is a popular Brain Computer Interface (BCI) method that offers hands-free control. Several applications were developed, including communication \cite{grau2014conscious}, driving a wheelchair \cite{tanaka2005electroencephalogram} and robotic arm control \cite{shedeed2013brain}. However, there are multiple challenges when using a BCI interface. The technology has long task completion time and high error rates \cite{albilali2013comparing}. BCI applications require high-level concentration and cognitive load which can lead to mental fatigue. A  specific cognitive state may be achieved in a quiet laboratory environment but is unlikely to be produced in the real world \cite{nicolas2012brain}. Overall, there is no consensus on what kind of skills are required to successfully drive a BCI controlled system \cite{curran2003learning}.

Eye-tracking provides a powerful alternative means of control for the disabled. Individuals with ALS or muscular dystrophy lose their muscle strength over time, eventually being unable to reach out and grasp. They also lose their ability to speak. However, they still have good control over their eyes \cite{kelly2013encyclopedia}. The gaze of a person can be interpreted as the direct output from the brain. Compared to detecting brain patterns using EEG, detection of eye movement is easier, faster and has higher accuracy \cite{nicolas2012brain}. The current state-of-the-art gaze-based assistive devices that are commercially available are mainly screen-based systems. Screen-based systems are useful for computer related tasks such as typing, sending email, browsing the web, as the user's gaze becomes the mouse pointer. By creating specific graphical user interfaces (GUIs), control of a system can be provided to the user. \textit{Eyedrivomatic}\footnotemark[3] uses arrows for users to fixate and move an electrical wheelchair. A drawback of screen-based systems is that they divert the user's attention from the outside world, essentially narrowing their vision. The ideal system should grant the user control by simply looking in the real world, in other words, the ability of free-viewing gaze control.

   \begin{figure}[tb]
      \centering
      \includegraphics[width=8cm]{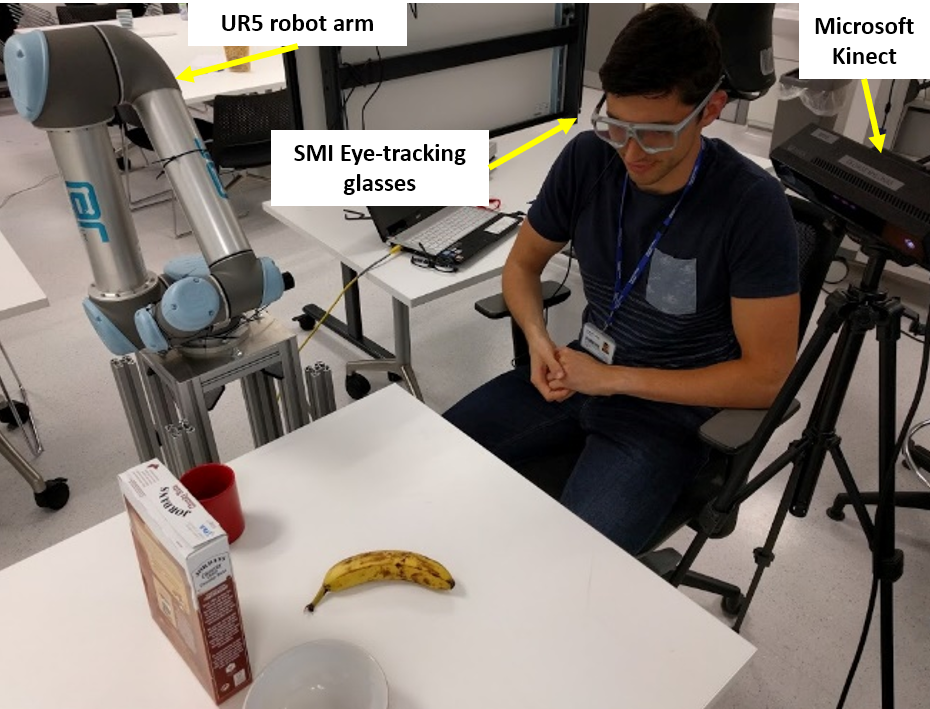}
      \caption{Setup of the proposed system.}
      \label{fig:setup}
   \end{figure}
   
\footnotetext[1]{Kinova Robotics: http://www.kinovarobotics.com/}
\footnotetext[2]{Exact Dynamics: http://www.exactdynamics.nl/}
\footnotetext[3]{Eyedrivomatic: http://www.eyedrivomatic.org/} 
In \cite{li20173d} the 3D point of regard is determined using ocular vergence, followed by neural networks to improve accuracy. Using 3D gaze the user can define the contour of a target object to be grasped by the robot. However, the lack of a world frame of reference restricts the capabilities of the system to predefined and calibrated spaces. Specifically, a long calibration procedure involving 64 calibration points is required, and a head stand to prohibit head movement.

The objective of this project is to develop a system that enables patients who suffer from motor impairment to gain independence in a free-view fashion (Fig. \ref{fig:setup}). We achieve this by integrating free-viewing 3D fixation localisation, automatic object recognition and trajectory planning into an assistive robotic system that performs activities of daily-living (ADL). This is done with the sole use of wireless eye-tracking glasses and one RGB-D camera. The user is offered two modes of interaction with objects in space using just eye-gaze as control input and a robotic arm for manipulation. In manual mode users can control the position of the robotic arm on their head's frame-of-reference. In automatic mode a pre-defined task associated with a gaze-selected object is executed. To the authors knowledge, this is the first system of its kind, providing unconstrained freedom and flexibility in unstructured environments.

\section{SYSTEM OVERVIEW}

\begin{figure}[b]
    \centering
    \begin{subfigure}[b]{0.18\textwidth}
        \includegraphics[width=\textwidth]{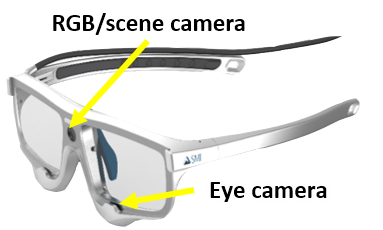}
        \caption{}
        \label{fig:etg}
    \end{subfigure}
 \qquad 
    \begin{subfigure}[b]{0.18\textwidth}
        \includegraphics[width=\textwidth]{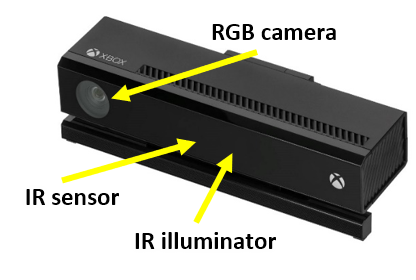}
        \caption{}
        \label{fig:kinect}
    \end{subfigure}
    \caption{(a) SMI eye-tracking glasses. (b) Microsoft Kinect v2 RGB-D sensor.}
\label{fig:equipment}
\end{figure}

The system consists of the following components:
\begin{itemize}
\item Eye-tracking glasses (ETG) from SensoMotoric Instruments (SMI) with an integrated scene camera with $1280 \times 960$ pixels resolution, as shown in Fig. \ref{fig:etg}. 
\item Microsoft Kinect v2 for RGB-D sensing (Fig. \ref{fig:kinect}), with full HD $1920 \times 1080$ pixels resolution at 30$Hz$ for its RGB camera and time-of-flight infrared depth sensor with 30$ms$ latency.
\item A 6 degrees of freedom (DoF) UR5 arm from Universal Robots, for manipulation. 
\end{itemize}

   \begin{figure}[t]
      \centering
      \includegraphics[width=7cm]{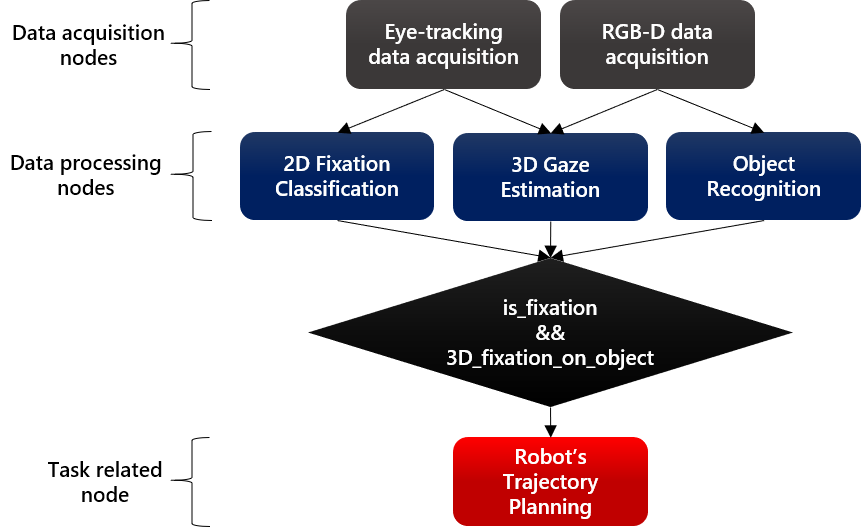}
      \caption{System Overview.}
      \label{fig:system}
   \end{figure}

The setup simulates a WMRM with a wheelchair-mounted RGB-D sensor and a user wearing the ETG.

To determine the user's visual attention, the point of regard (PoR) in 3D space must be determined first. A high-level description of this task involves the following steps: (1) The RGB image information from the ETG's scene camera and the Kinect colour and depth camera images are used to estimate the ETG pose. (2) Once this pose is retrieved, the 3D PoR can be computed as the intersection between the gaze vector and the 3D reconstructed scene. (3) Objects that are in front of the user are identified and their pose estimated. (4) Once the 3D fixation point lies on the object, the UR5 arm executes a task associated with the chosen object, depending on the mode selected. Fig. \ref{fig:system} shows the structure of the system. Object grasping is not dealt with for this project. Instead, an end-effector with a magnet attachment is used to "grip" objects.

\section{METHODOLOGY}

The system is developed in \textit{Robot Operating System (ROS)} with C++. \textit{ROS}, being the middleware, allows effective communication to be set up between all the elements of the system. For the current implementation, a Windows 7 computer is used for acquiring and streaming the ETG data and a Linux PC with Ubuntu 14.04 is used for all other modules. The Linux PC runs on Intel Xeon Processor, NVIDIA GTX 1050 2GB, 16 GB RAM. This section discusses the methodology behind the core modules of the system, namely the coordinate frames registration, 2D fixation classification, head pose estimation, 3D gaze estimation, object recognition, trajectory planning and operation modes.

\subsection{Coordinate Frames Registration}
In the proposed system, we use the robot's coordinate system as the world frame of reference. To align multiple local frames to the global one, calibration between the robot and the RGB-D camera is necessary. The method we chose involves manually positioning the robot's end effector on the corners of a printed checkerboard, which is visible by the RGB-D camera at the same time. By performing this we estimate the rigid transformation between the robot and the RGB-D camera, as both are assumed rigidly mounted on the frame of a wheelchair. The transformations shown in (Fig. \ref{fig:transform}) are described by the following equations: 

   \begin{figure}[bt]
      \centering
      \includegraphics[width=7cm]{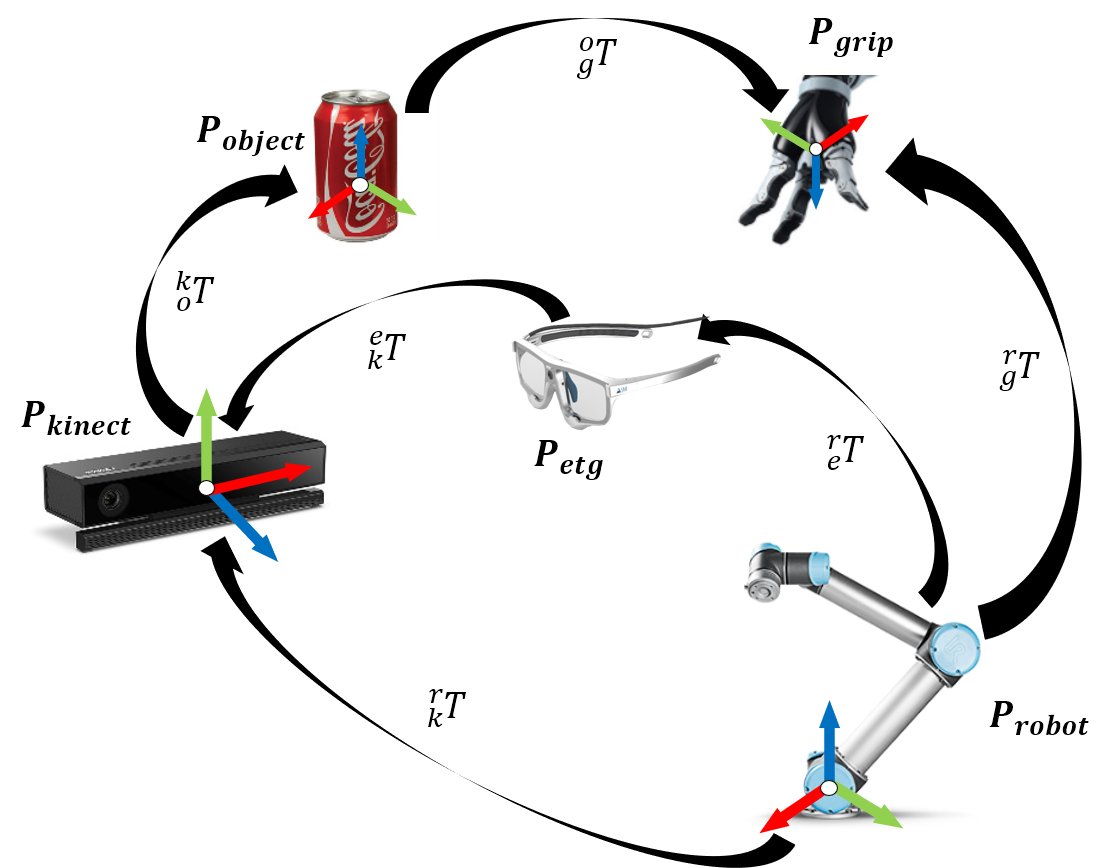}
      \caption{The transformations among the coordinate systems.}
      \label{fig:transform}
   \end{figure}

\begin{eqnarray}
\label{trans1}
_{g}^{r}\textrm{T} = _{k}^{r}\textrm{T} * _{o}^{k}\textrm{T} * _{g}^{o}\textrm{T}\\
\label{trans2}
_{k}^{r}\textrm{T} = _{e}^{r}\textrm{T} * _{k}^{e}\textrm{T}
\end{eqnarray}

\subsection{2D Fixation Classification}
The ETG provide the 2D PoR on the user's head frame-of-reference. As a safety precaution, the activation routine of the robotic manipulation task is based on the 2D fixation dwell time. First, the velocity of eye movement is estimated \cite{mould2012simple} and a threshold of 36$deg/s$ is set to filter out fast saccadic movement. Moreover, we only consider fixations over a dwell time threshold of 2$s$.

\subsection{Head Pose and 3D Gaze Estimation}

\begin{figure}[tb]
      \centering
      \includegraphics[width=6cm]{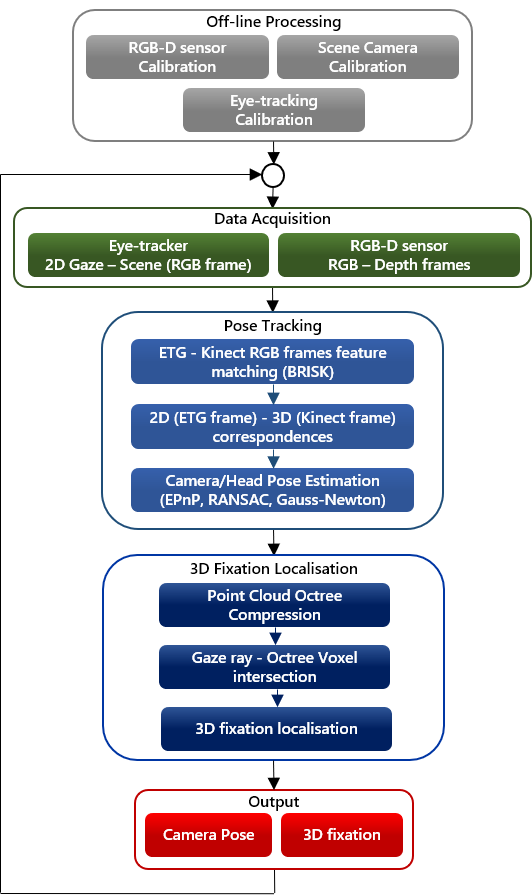}
      \caption{3D gaze estimation module.}
      \label{fig:gaze3d}
   \end{figure}

The 3D gaze estimation component is based on the novel framework proposed in \cite{kogkas2017gaze} and relies on the combination of advanced computer vision techniques, RGB-D cameras and ETG. With reference to  Fig. \ref{fig:gaze3d} the process consists of two tasks: user's head pose estimation and 2D to 3D gaze mapping. The user's head pose is equivalent to the ETG's RGB/scene camera pose in space. For the camera pose estimation, BRISK features \cite{leutenegger2011brisk} are detected and matched in both the ETG's frame and the RGB camera frame of the RGB-D sensor. The RGB-D extrinsic camera calibration \cite{iai_kinect2} provides the depth values of the matched RGB features and consequently the 2D-3D correspondences for the ETG's features (2D points on ETG's RGB/scene camera -- respective 3D coordinates in the Kinect's coordinate system). Then, EPnP with RANSAC and Gauss-Newton Optimisation \cite{lepetit2009epnp} provide the ETG's scene camera pose in space. For the last step, we use ray tracing to backproject the gaze ray from the compressed model of the 3D reconstructed environment (to improve performance) on the estimated camera pose origin, allowing real-time and free-viewing 3D fixation localisation. Fig. \ref{fig:gaze3d} outlines the 3D gaze estimation framework.

\subsection{Object Recognition and Selection}
For object detection and pose estimation, \textit{LINEMOD} \cite{hinterstoisser2011multimodal} was used with \textit{Object Recognition Kitchen (ORK)} \cite{ork_ros} as a backend. \textit{LINEMOD} is a real-time template matching method and \textit{ORK} is a framework which offers various techniques for object recognition. This includes setting up a local database to store a 3D mesh file of each object and generating the templates of the stored objects.

The 3D fixation corresponds to a point from the point cloud of the Kinect scene. To identify whether this point is on any of the recognised objects, a set of neighbouring points around the fixations is compared with \textit{ORK}'s point cloud. A \textit{k-dimensional (k-d)} tree algorithm was deployed to search for nearest neighbours with a radius of 1$cm$. In case the 3D fixation is detected within the point cloud, the next step is to identify the specific object being fixated. As \textit{ORK} provides the centroid for each object, the Euclidean distances between the fixation point and the centroids for all objects were calculated. The object with the shortest distance would be the fixated object and its pose then becomes the input for the trajectory planning module. Obstacle detection has yet to be implemented at this stage.

\subsection{Trajectory Planning}
To control the UR5 arm, the \textit{Moveit!} framework \cite{sucan2013moveit} was selected. \textit{Moveit!} is an open-source software for robotic manipulation, motion planning and control and is fully integrated with \textit{ROS}. Therefore, it allows easy communication with our Kinect perception and gaze-control module. 

From the object recognition module, the pose of the selected object's centroid is received. From the object's centroid, the contact point for the magnetic gripper is calculated on the object's surface, based on its known dimensions. Depending on which objects are selected, different manipulation poses are established for the task intended (pre-grip poses), followed by object pick up. All movements in the manipulation module can be divided into two types: \textit{motion planning} and \textit{cartesian path planning}. Motion planning is based on planning a collision-free path from the current state to a designated pose, while cartesian path planning relies on computation of waypoints. The former was used to generate a trajectory from the robot's home pose to the object's pre-grip pose, while the latter was deployed once the arm reached the pre-grip position and in the \textit{manual mode} (\ref{sssec:manmod}). Safe zones, such as where the user is and the table, have been set up to prevent path planning from taking place within this space. 

\subsection{Operation Modes}
The system offers the user two modes of interaction with the objects. 

   \begin{figure}[tb]
      \centering
      \includegraphics[width=8.8cm]{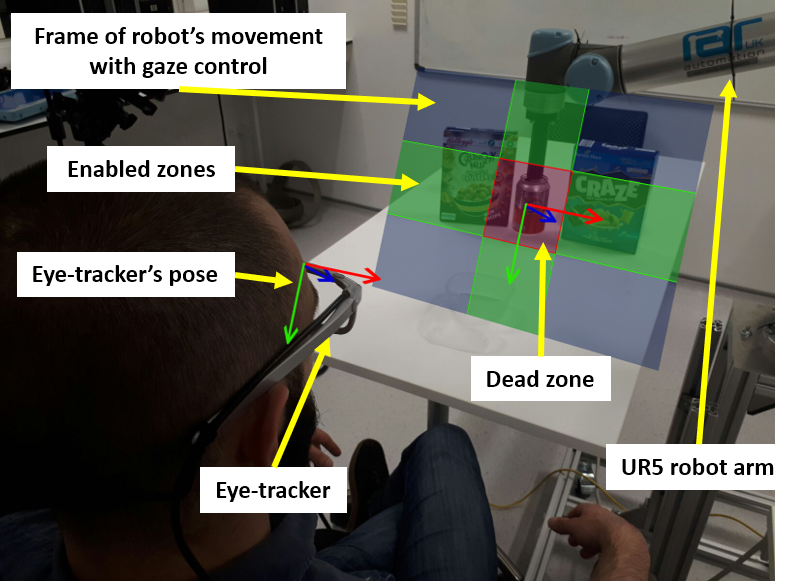}
      \caption{Control plane corresponding to the user's view in manual mode.}
      \label{fig:manmod}
   \end{figure}

\subsubsection{Automatic Mode}
The automatic mode executes a pre-defined task associated with a selected object. The user triggers the task by fixating on a recognised object. According to \cite{chung2013functional}, meal preparation and drink retrieval were considered top desired tasks for disabled patients. It was decided that the automatic mode should incorporate these functions.

\subsubsection{Manual mode} \label{sssec:manmod}
The manual mode provides the user with positional control of the end-effector in the X, Y, Z axes with respect to the ETG frame. The transformation between the ETG and the world frame, which is aligned to the robot frame, is initially calculated (\ref{trans2}). This allows the end-effector's position to be determined in the ETG frame. The 2D gaze coordinates from the ETG are translated to a movement in one of the three directional axes. A dead zone of 300$\times$300 pixels was created in the centre of the ETG RGB image. The robot will not move if the 2D PoR is within this zone. If the user's PoR is to the left of this zone, the robot moves to the left by a small pre-defined offset of 2$cm$; this also applies to right, up and down. In and out depth movement is performed by closing one or the other eye. This discrete motion of the manipulator was chosen over continuous action, as it was found that the user can perform the task safer and more intuitively. Orientation control is not included at this stage as this might increase complexity for the user. Once the new pose has been determined in the ETG frame, this gets transformed into a coordinate in the robot frame and the robot moves in a step manner. Fig. \ref{fig:manmod} shows a visualisation of the control plane projected in front of the user, in the same orientation as the ETG pose (scene camera). Synthesised voice feedback acknowledging the directional commands is provided for assistance, as the user's centre of gaze may not always be on the end-effector and also it was found that feedback helps with the overal confidence of the user during task execution. The small steps allow the user to perform fine positioning of the end-effector, ideal for situations where the pose of an object is inaccurately determined due to point cloud distortions or other artifacts.

\subsection{Application Workflow}
The workflow of the system starts with an off-line pipeline required by the object recognition module and the Kinect-to-Robot registration. First, the 3D mesh models of the objects are loaded to \textit{ORK}. Then, the RGB-D camera is registered to the UR5 robot (world coordinate system). Finally, the user wears the ETG and performs a standard eye-tracking calibration procedure to align the ETG's scene camera frame with captured gaze vectors while fixating on three different and spread out in space points. Finally, the user is ready to fixate on the trained objects to trigger the automatic or the manual mode.

\section{Experimental Evaluation}

\subsection{3D Gaze Estimation Evaluation}
The accuracy and computation time of the 3D fixation localisation were examined. A subject was recruited and asked to fixate on 10 predefined targets from 6 different positions. The estimated 3D fixations were compared to the actual ones by measuring their Euclidean distance. Moreover, the time interval between the moments the subject's PoR was classified as a fixation and the 3D fixation was computed.

\subsection{Trajectory Planning Performance}
The success rate of the trajectory planning was examined. On the Kinect cloud, 3D points which belong to the objects of the experimental setup were manually selected and the rate of successful trajectory planning was estimated. The time between the moment a point was selected and the moment the robot started the object-specific task was also measured. Two objects were used for this experiment, a mug and a cereal box, which require different griping orientation by the robot. Each was placed in 3 different positions on a table, within the robot's maximum reach. The process was repeated 10 times for each object.

\subsection{Overall Evaluation of the System}
An experimental study was performed to assess the usability of the overall system. Two experiments were carried out to validate each operation mode. The study measured the system's performance objectively as well as the users' subjective experience. The experiments were carried out in a well-lit room and objects were placed on a nonreflective table. Five healthy subjects, aged between 21--26 years participated in the study. Two subjects had normal vision while the rest had corrected vision. Prior to the experiment, each subject was briefed on the purpose of the study, the technology involved and the expected tasks outlined below. A three-point calibration was performed at the beginning of each experimental session to ensure that the ETG were correctly tracking the subject's pupils and subsequently providing the accurate gaze direction.

\subsubsection{Automatic Mode}
The experimental setup involved placing a coffee mug, a cereal box, a bowl, a banana and a plastic container on a table. Fig. \ref{fig:expsetup} shows the setup of the experiment. All objects were placed between 100--120$cm$ away from the Kinect sensor but within the UR5's working space (85$cm$ reach). Three tasks were implemented for the study:

   \begin{figure}[tb]
      \centering
      \includegraphics[width=8.8cm]{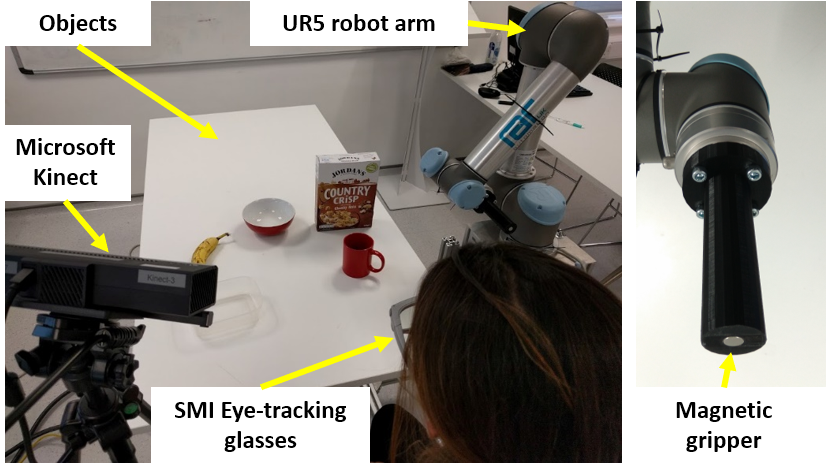}
      \caption{Experimental setup simulating a WMRM, assuming an external mount on the left side of the wheelchair for the Kinect sensor.}
      \label{fig:expsetup}
   \end{figure}
   
\begin{itemize}
\item By fixating on the mug, the robot would reach inside the mug and bring it towards the user. 
\item By fixating on the cereal box, the robot would pick it up, locate the bowl and pour cereals into it. The robot then places the box beside the bowl.
\item Fixating on the bowl, the banana and the plastic container should not prompt any robotic action (the latter two are not loaded to \textit{ORK} and are considered distractors). 
\end{itemize}

An instructor then requests the subject to fixate on one of the objects on the table to prompt the above tasks. The order of fixation was given randomly by the instructor. Once the set of fixations on five different objects has been completed, the positions of the objects were randomised for the next set. Each subject was asked to perform three sets of trials.

\subsubsection{Manual Mode}
The object of choice for this experiment is an aluminium soft drink can. The reason being, reflective objects do not get accurately detected by the RGB-D sensor due to multipath interference, therefore the estimated pose is incorrect. We made use of this occurrence and requested the subjects to fixate on the can. The system would output the incorrect pose of the object and the robot would move towards the pre-grip pose, somewhere close to the can. The subjects were then instructed to steer the robot with their gaze to pick up the can and place it in a plastic container with dimensions 12$\times$15$\times$5$cm$ positioned 30$cm$ away from the can. The subjects were instructed to activate each direction once with the instructed eyes gestures prior to the experiment, but no training runs were provided. This experiment was performed twice for every subject.

\subsubsection{Control Modalities Evaluation}
Individual elements were evaluated simultaneously during the study along with the overall success rate of the system. Measurements for automatic and manual mode are as follows:

\begin{itemize}
\item\textit{{Automatic Mode}}

\textit{Successful selection of the object} -- Five different objects were used in the experiment to assess the performance of the object recognition and 3D gaze estimation elements of the system. It was considered a success when the system planned a path to the predefined pose of the selected object.

\textit{Activation time} -- The elapsed time from when the user begins fixating on the object to when the robot starts moving. This outcome signifies real-time usability.

\textit{Task completion success rate} -- When the robot successfully performs the intended task that corresponds to the object selected, without colliding with other objects or faulting out.

\item\textit{{Manual Mode}}

\textit{Task completion time} -- The time elapsed from the user gaining control of the robot to when the can touched the bottom of the container.

\textit{Task completion success rate} -- Successful or not successful.

\end{itemize}

Selection of object and activation time were not measured in manual mode as this was validated during automatic mode. After the experiment, the subjects were asked to fill out a questionnaire regarding their experience using the assistive system. A 5-point \textit{Likert scale} ranging from 1 -- strongly disagree to 5 -- strongly agree, was provided to rate their opinion.

\section{Results and Discussion}

\begin{figure*}[t]
\begin{center}
\begin{minipage}{0.14\linewidth}
  \centering
\centerline{\includegraphics[width=\textwidth, height=5.5cm]{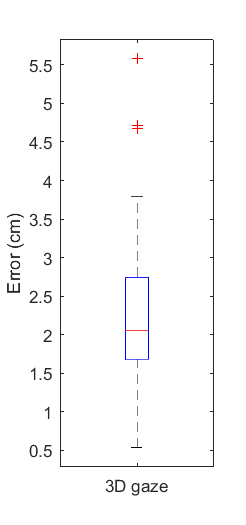}}
  \centering {(a)}
\end{minipage}
~
\begin{minipage}{0.32\linewidth}
  \centering
 \centerline{\includegraphics[width=\textwidth, height=5.5cm]{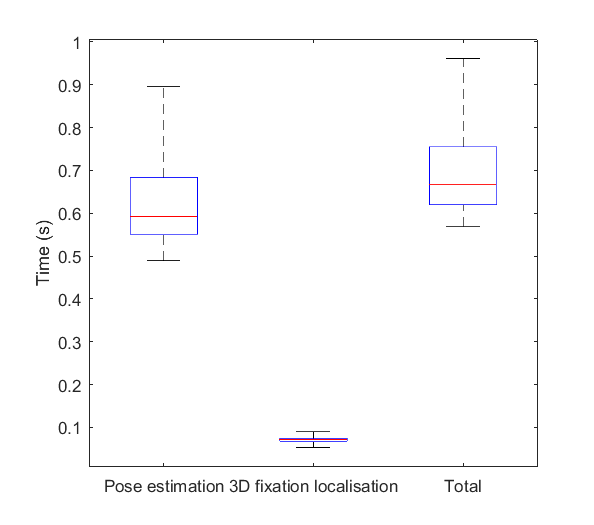}}
  \centering {(b)}
\end{minipage}
~
\begin{minipage}{0.22\linewidth}
  \centering
\centerline{\includegraphics[width=\textwidth, height=5.5cm]{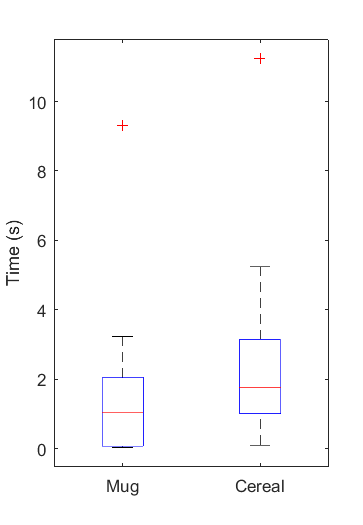}}
  \centering {(c)}
\end{minipage}
~
\begin{minipage}{0.22\linewidth}
  \centering
 \centerline{\includegraphics[width=\textwidth, height=5.5cm]{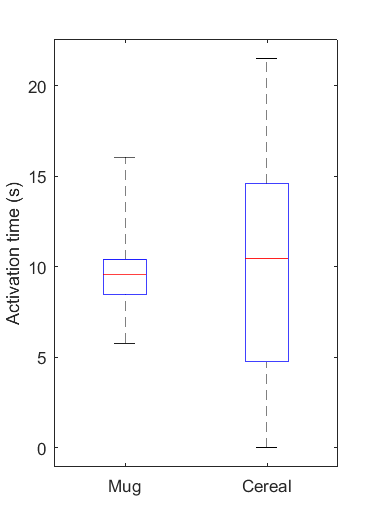}}
  \centering {(d)}
\end{minipage}
\end{center}

    \caption{(a) 3D gaze error and (b) time of pose estimation and 3D fixation localisation. (c) Path planning time of the robot, targeting the cereal and the mug. (d) Activation times for mug and cereal, from user beginning fixating to robot moving. This timing includes the 2$s$ dwell time threshold and the 1$s$ of \textit{ROS} sleep.}

\label{fig:plots}
\end{figure*}

\begin{figure*}[t]
\begin{center}
\begin{minipage}{0.45\linewidth}
  \centering
\centerline{\includegraphics[width=7cm]{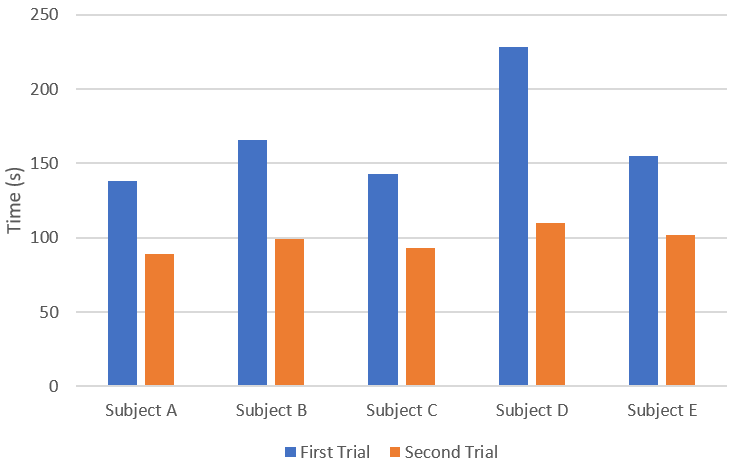}}
  \centering {(a)}
\end{minipage}
~
\begin{minipage}{0.45\linewidth}
  \centering
 \centerline{\includegraphics[width=7cm]{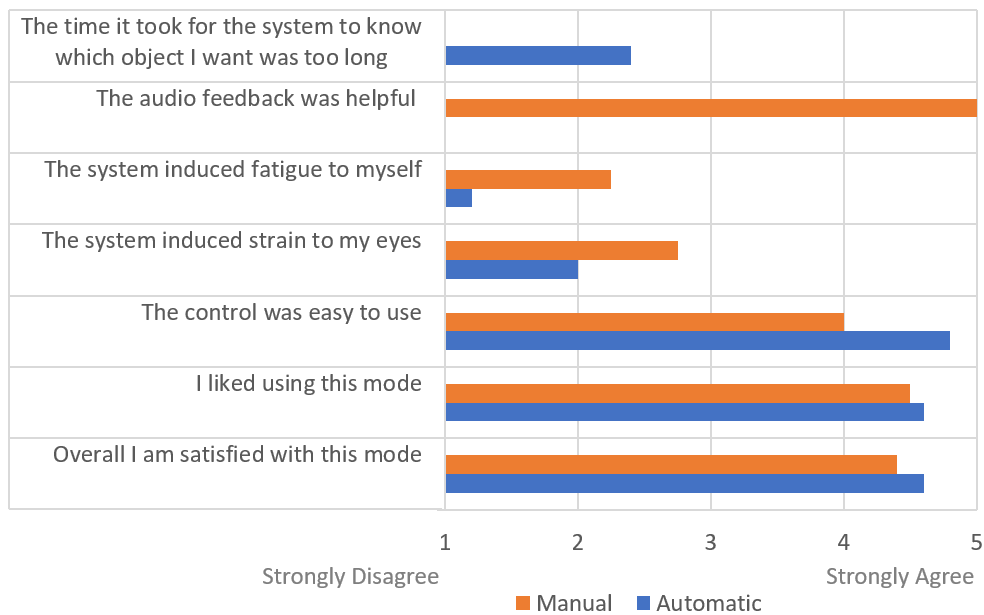}}
  \centering {(b)}
\end{minipage}
\end{center}

    \caption{(a) Completion time for each subject for pick and place task. (b) Subjects' feedback for both manual and automatic modes.}

\label{fig:graphs}
\end{figure*}

\subsection{3D Gaze Estimation Evaluation}
The 3D gaze estimation is evaluated in terms of accuracy and computational time. For this, 10 markers were placed on objects positioned at different depths. The distance between the RGB-D camera and the objects is 100--130$cm$, which forms a realistic workspace for the specific application, considering the UR5's maximum reach of 85$cm$. The average error is 2.31$\pm$1.03$cm$ and the computation time was measured at 0.69$\pm$0.09$s$ (Fig. \ref{fig:plots}a-b). The computation time comprises of the camera pose estimation and the 3D fixation localisation parts.

\subsection{Trajectory Planning Performance}
The activation time of the robot's path planning was measured. As shown in Fig. \ref{fig:plots}c, the interval is 2.3$\pm$2.26$s$ for the cereal and 1.23$\pm$1.81$s$ for the mug. Moreover, 100\% success rate was achieved by the trajectory planning modules, while 91.67\% was the rate for the successful grasping of the targeted objects.

\subsection{Overall Evaluation of the System}
\subsubsection{Automatic Mode}
Table \ref{table_I} shows the success rate of the system modules along with the overall success rate for the automatic mode. The high success rate of the gaze-guided object recognition demonstrates that the system is capable of recognising the objects on the table and the 3D gaze estimation is accurate enough to trigger the intended robotic task. The path planning can also be considered reliable, failing only one time out of the 30 attempted plans. The overall system success rate dropped below 90\%, despite the previous modules having over 96\% success rate. This is due to the non-deterministic nature of the sampling-based motion planner. Although the generated path was valid, without obstacle detection implemented it was possible that it collided with an object as it travelled through its trajectory. This, however, was considered a fail during the experiment.

\begin{table}[b]
\caption{Automatic Mode Success Rates}
\label{table_I}
\begin{center}
\begin{tabular}{|c||c|}
\hline
Gaze Guided Object Recognition & 98.67\%\\
\hline
Path Planning & 96.67\%\\
\hline
Overall System & 86.67\%\\
\hline
\end{tabular}
\end{center}
\end{table}

Fig. \ref{fig:plots}d shows the activation times for each object. The resulting average activation time was 9.92$\pm$4.78$s$. Removing the fixation requirement of 2$s$ and the \textit{ROS} node sleep rate of 1$s$, the average time to determine the user's 3D fixation point and to plan a valid path is 6.92$s$. Although activation time is an important aspect for a Human-Robot Interaction system, studies showed that patients did not feel the time to complete the task was significant, but rather they are content with being able to perform the task independently \cite{chung2013functional}.

\subsubsection{Manual Mode}
All subjects were able to complete the task of picking up the can and placing it in the plastic container, demonstrating a success rate of 100\%. Each subject showed the ability to grasp the control within the first run and improved the execution speed on the second run, as seen in Fig. \ref{fig:graphs}a. This study showed that the system was intuitive enough as no training was provided beforehand.

\subsubsection{User Experience}
All subjects' feedback is shown in Fig. \ref{fig:graphs}b. Questions regarding the negative aspects of the system generally received a low score, indicating the users were not frustrated or fatigued while operating the system. The time for the system to know which object was targeted trended towards a neutral score. This is related to the activation time and how some users experienced a longer wait in some occasions. The cause could arise from the inability to detect their eyes, the inability to compute the ETG pose or the random nature of motion planning. The question related to the system inducing strain to the user's eyes for the manual mode had a neutral score of 2.75. This was expected as the person is fully controlling the robot compared to the other mode, which is relying on activation just by the fixation. The positive aspects of the system received high scores, with the overall satisfaction score being 4.6 / 5.

\subsection{System Limitations}
Being at an early stage of development, the system has some limitations, which can affect its success rate and practical usability. As mentioned previously, the current implementation does not yet include obstacle detection, therefore the valid paths that trajectory planning produces have the possibility of objects collisions. Overcoming this limitation is feasible by using Octomap \cite{hornung2013octomap} to convert the RGB-D data into occupied space. \textit{Moveit!} will then be able to plan around the occupied region and generate collision-free trajectories.

In order for the system to be usable in everyday life, there is the evident need of a grasper. Integration to a commercial WMRM solves this issue and the product also contains pre-defined ADL tasks. However, prior to integration, the system needs to be able to switch between the different modes during runtime. This allows the patient to correct for any errors the system makes in pose estimation while granting them total control of the manipulation. Potential methods for switching between modes can range from closing one's eyes for a certain duration, draw a pattern with gaze gesture or even using additional hardware, such as Augmented Reality (AR) glasses, just to name a few possibilities. Finally, the last mile, i.e. allowing the robotic manipulator to approach the user's lips and complete the task, is not handled with the current version of the system, but this can be solved with an additional camera for face tracking.

\section{Conclusion}
In this paper, we presented a proof-of-concept for a gaze guided assistive robotic system used in a real environment. The system relies on wireless eye-tracking glasses and an RGB-D camera to achieve free viewing 3D gaze estimation in real-time, object recognition and trajectory planning. A robotic arm is used to execute activities of daily living, such as meal preparation and drink retrieval. Automatic and manual operation modes were implemented to provide useful interaction between the user and desired objects. The results show that the system is accurate, intuitive and easy to use even without training. For its practical deployment and extensive evaluation with actual patients, collision avoidance will have to be implemented and the RGB-D camera and a lightweight robotic arm have to be integrated with a wheelchair.

As the system is designed for home use, 3D models of household items can be added to the object recognition database. We can utilise the RGB-D sensor to scan the object and create a 3D mesh of it. This will allow the patient to scan objects of their choice, creating a personalised database.

Additional hardware, such as AR glasses, will enhance the user experience and allow further independence to the user, bringing the system closer to its actual integration in the everyday life of patients with severe motion disabilities.

Further work will involve actual patients. 

\bibliographystyle{IEEEtran}								%
\bibliography{bib}		

\end{document}